# StackLiverNet: A Novel Stacked Ensemble Model for Accurate and Interpretable Liver Disease Detection

1st Md. Ehsanul Haque
*Dept. of CSE*
*East West University*
Dhaka, Bangladesh
ehsanulhaquesohan758@gmail.com

2nd S. M. Jahidul Islam
*Dept. of CSM*
*Bangladesh Agricultural University*
Mymensingh, Bangladesh
jahidul.ict@bau.edu.bd

3rd Shakil Mia
*Dept. of CSE*
*Daffodil International University*
Dhaka, Bangladesh
shakilmia90923@gmail.com

4th Rumana Sharmin
*Department of Food and Nutrition*
*University of Dhaka*
Dhaka, Bangladesh
rumanasharminshorna@gmail.com

5th Ashikuzzaman
*Dept. of CSE*
*University of Barishal*
Barishal, Bangladesh
ashikuzzaman.cse7.bu@gmail.com

6th Md Samir Morshed
*Department of Botany*
*Cumilla Victoria Government College*
Cumilla, Bangladesh
samirmorshed85@gmail.com

7th Md. Tahmidul Huque
*Dept. of CSE*
*Bangladesh University of Business and Technology*
Dhaka, Bangladesh
md.tahmidulhuque@gmail.com

*Abstract*—Liver diseases are a serious health concern in the world, which requires precise and timely diagnosis to enhance the survival chances of patients. The current literature implemented numerous machine learning and deep learning models to classify liver diseases, but most of them had some issues like high misclassification error, poor interpretability, prohibitive computational expense, and lack of good preprocessing strategies. In order to address these drawbacks, we introduced StackLiverNet in this study; an interpretable stacked ensemble model tailored to the liver disease detection task. The framework uses advanced data preprocessing and feature selection technique to increase model robustness and predictive ability. Random undersampling is performed to deal with class imbalance and make the training balanced. StackLiverNet is an ensemble of several hyperparameter-optimized base classifiers, whose complementary advantages are used through a LightGBM meta-model. The provided model demonstrates excellent performance, with the testing accuracy of 99.89%, Cohen Kappa of 0.9974, and AUC of 0.9993, having only 5 misclassifications, and efficient training and inference speeds that are amenable to clinical practice (training time 4.2783 seconds, inference time 0.1106 seconds). Besides, Local Interpretable Model-Agnostic Explanations (LIME) are applied to generate transparent explanations of individual predictions, revealing high concentrations of Alkaline Phosphatase and moderate SGOT as important observations of liver disease. Also, SHAP was used to rank features by their global contribution to predictions, while the Morris method confirmed the most influential features through sensitivity analysis. These findings prove that StackLiverNet is a precise, viable, and explainable model that may support clinicians in the early diagnosis of liver disease and enhance patient care.

*Index Terms*—Liver disease classification, StackLiverNet, interpretability, explainable AI (XAI), medical diagnosis.

## I. INTRODUCTION

Liver diseases are one of the largest health-related issues worldwide, as they cause more than 2 million deaths every year, and two of the most widespread causes of liver diseases are cirrhosis and liver cancer [1]. Clinical evolution of liver disease is usually silent in the initial phases of the disease and it is imperative that the diagnosis is made in a timely and accurate manner in order to enhance the outcome in patients and to ameliorate the healthcare cost which has been shown to be high in such patients [2]. However, current diagnostic models used in computational medicine have several problems of unbalanced data distribution, irrelevant or redundant clinical features, as well as the lack of interpretability, which jointly limit their applicability in practical health care. The limitation of traditional classifiers is that they might not be able to capture the subtle pattern in the high-dimensional clinical data, resulting in poor performance and high misclassification, especially when the disease classes are minorities [3] [4]. In order to overcome those drawbacks, we introduce in this paper, StackLiverNet, an explainable stacked ensemble model that aims to enhance the predictive accuracy and stability of liver disease classification. The framework contains a full preprocessing pipeline, containing outlier management, feature scaling, and dimensionality decrease with recursive feature elimination with cross-validation (RFE-CV). Random undersampling is used to alleviate the class imbalance problem, and stacking ensemble strategy with multiple base learners with optimized hyperparameters is used. The final meta-classifier integrates the strengths of individual models, enabling robust generalization and improved predictive accuracy.

In addition, the decisions of the models are explained by making them transparent, by adding local interpretable model-agnostic explanations (LIME), allowing clinicians to see how each feature contributes to each prediction, which is essential in healthcare use cases.

The main contributions of this study are as follows:
1) Proposal of StackLiverNet, a novel stacking-based ensemble for liver disease detection.
2) In-depth preprocessing pipeline to enhance data quality, incorporating RFE-CV for optimal feature selection.
3) Reduction of misclassification errors and enhancement of predictive accuracy through ensemble learning and hyperparameter-optimized base classifiers, along with effective class imbalance handling.
4) Minimization of computational cost while maintaining model performance.
5) Integration of explainable AI techniques (LIME, SHAP, and Morris method) to ensure both local and global interpretability, thereby enhancing transparency and supporting clinical decision-making.

This paper has shown that StackLiverNet outperforms conventional methods, not only by classification measures but also by the transparency of the model, thereby showing potential to become a useful tool in the initial screening of liver disease.

## II. RELATED WORK

Several recent studies have explored machine learning and deep learning approaches for liver disease detection using publicly available datasets, notably the Kaggle Liver Disease Patient Dataset (LDPD).

Ganie et al. [5] have compared the ensemble learning methods such as boosting, bagging and voting with various classifiers. Their Gradient Boosting model had great accuracy (98.80%) and excellent precision and recall (98.50%), but simple mean/median imputation and SMOTE oversampling could be quite biasing and not very interpretable.

Hendi et al. [6] developed a CNN-LSTM hybrid model on liver disease subtyping achieving an accuracy of 98.73%. Their deep learning method achieved better results than standalone models but lacked transparency that is common with deep learning methods; they also did not explicitly tackle the issue of missing data.

Khatun et al. [7] proposed an ensemble method Tree-selection and Stacked Random Forest (TSRF) and reported 99.72% accuracy on Kaggle dataset. To make their models more transparent, they included explainability techniques like SHAP and LIME. They are, however, dependent on small dataset, which brings into question generalizability and real-world validation.

Jena et al. [8] evaluated boosting algorithms, such as Gradient Boosting and CatBoost on several datasets with a focus on hyperparameter search and cross-validation. Their finest model achieved 98.80% accuracy, and the research did not focus on data cleaning or explainable AI, which reduces their use in clinics.

Nilofer et al. [9] tested Random Forest, XGBoost against an intrinsically interpretable Explainable Boosting Machine (EBM). Despite the fact that EBM showed high accuracy (nearly 99.8%) and some more transparency, no detailed information on data preprocessing and sensitivity was implemented in the study.

Ruhul Amin et al. [10] proposed a unified machine learning framework on advanced feature extraction using chronic liver disease classification. They used the dimensionality reduction techniques such as Principal Component Analysis (PCA), Factor Analysis (FA), and Linear Discriminant Analysis (LDA) in order to improve the quality of extracted features. Their model had an accuracy of 88.10% and AUC of 0.8820, which showed promising results as a stable diagnostic tool to be used by clinicians and pathologists.

Overall, these studies have shown promising findings on the prediction of liver disease using different types of ensemble and machine learning models, but the general drawbacks of these models are the lack of processing of missing or imbalanced data, low interpretability, and the absence of external clinical validation. The proposed StackLiverNet framework fills these gaps since it incorporates state-of-the-art preprocessing, stacking ensembles, and explainable AI to enhance accuracy, reliability, and explainability.

## III. METHODOLOGY

In this section, we provide a detailed explanation of the research methodology. The overall workflow of the proposed approach is illustrated in Figure 1, which outlines the process from data acquisition to ensemble-based liver disease detection and subsequent model interpretation techniques.

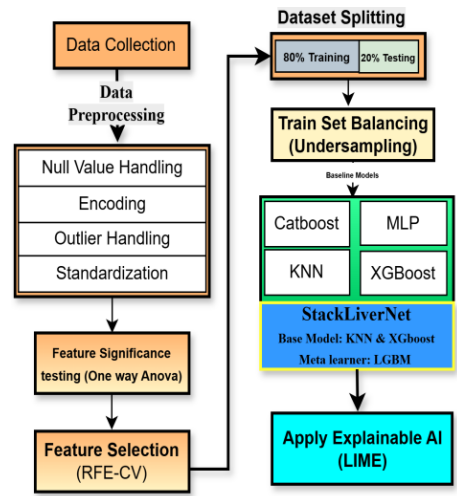

Fig. 1: The proposed workflow for Liver Disease detection.

### A. Data Collection

The dataset was obtained from Kaggle and contains a total of 26,303 samples [11]. Among these, 16,749 correspond to liver patients, while the remaining represent non-liver cases. The dataset comprises 11 features, including one target variable indicating the presence or absence of liver disease. Among the features, one is categorical, one is boolean, and the remaining nine are numerical predictors.

### B. Data Processing Pipeline

There is one categorical variable, *Gender*, in the dataset, where missing values were imputed as mode. The imputed data was then encoded to categorical values to suit the model. In other features, missing values were removed by simply removing the rows to preserve the data integrity. The interquartile range (IQR) was used to detect outliers and Winsorization was applied to constrain the effect of extreme values. The target variable, *Result*, was recoded with 0 and 1 replacing original values 1 and 2 respectively to indicate the binary classification task. Lastly, the standardization of all the numerical features was done to guarantee equal predictor scaling.

### C. Feature Significance Testing

A one-way ANOVA was performed to determine the significance of each feature against the target variable where a significance level of 0.05 was used (95% confidence) [12]. The p-value of the *Gender* feature was 0.5636 and that of the *Age* feature was 0.5113 which is above the threshold and thus, they do not exhibit any significant correlation. As a result, these two characteristics were eliminated as dataset to be further analyzed.

### D. Feature Selection

The Recursive Feature Elimination with Cross-Validation (RFE-CV) was used to determine the most pertinent features to be used by the model [13]. The strategy was assessed through testing tiny subsets of 3, 5, and 7 features using cross-validation (CV) scores. The best CV performance was obtained with both 5 and 7 features, but 5 features were finally chosen to simplify computations with preserving high predictive accuracy. The importance of the selected feature is demonstrated in Table I whereas Figure 2 demonstrates the CV scores through the RFE process. In addition, the pseudocode Algorithm 1 presents a step-by-step process of feature selection in detail.

TABLE I: Selected Features and Their Importance Scores

| Feature No. | Feature | Importance Score |
|---|---|---|
| 1 | Total Bilirubin | 0.2015 |
| 2 | Alkaline Phosphatase | 0.2308 |
| 3 | Aspartate Aminotransferase | 0.2078 |
| 4 | Alanine Aminotransferase | 0.1919 |
| 5 | Albumin | 0.1680 |

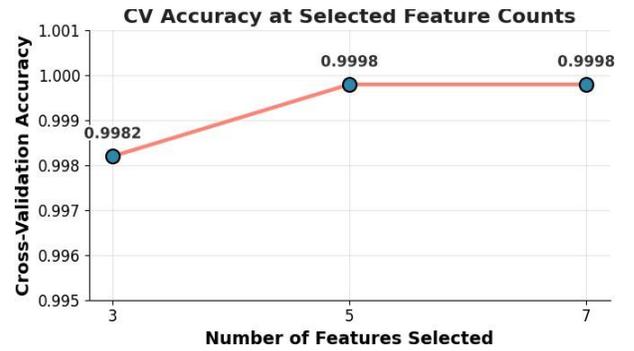

Fig. 2: Cross-validation scores for different numbers of features during Recursive Feature Elimination with Cross-Validation (RFE-CV).

---

**Algorithm 1** RFECV with Random Forest

**Require:** Features *X*, labels *y*, folds *k* = 5
**Ensure:** Optimal features F*, accuracy list **A**, accuracies at 3, 5, 7 features
 0: F ← X; **A** ← []; **S** ← {}
 0: **while** |F| > 1 **do**
 0:  Train RF with *k*-fold CV on F, get mean accuracy *a*
 0:  Append *a* to **A**; if |F| ∈ {3, 5, 7}, set **S**[|F|] ← *a*
 0:  Remove least important feature
 0: **end while**
 0: F* ← features at highest accuracy in **A**
 0: **return** F*, **A**, **S** =0

---

### E. Dataset spliting and balancing

The dataset was split into training and testing sets using an 80-20 ratio. The original training set exhibited class imbalance, with significantly more samples in Class 0 ( Disease) than Class 1 (No Disease). To address this, undersampling was applied to the majority class to balance the training data. Table II summarizes the class distribution across the original training set, the balanced training set after undersampling, and the test set.

TABLE II: Class distribution before and after undersampling

| Dataset | 0 (Disease) | 1 (No Disease) |
|---|---|---|
| Original Training Set | 13,414 | 5,282 |
| Training Set (After Undersampling) | 5,282 | 5,282 |
| Test Set | 3,335 | 1,339 |

### F. Baseline Models

In this work, four baseline classifiers are used to compare the predictive accuracy on liver disease dataset: CatBoost, Multi-Layer Perceptron (MLP), k-Nearest Neighbors (KNN), and XGBoost. CatBoost is a gradient boosting algorithm that is said to effectively deal with categorical features and discourage overfitting by means of ordered boosting. The MLP is a feedforward neural network which has the ability to capture non-linear complex patterns. KNN is a distance-based data point classification model that uses the concept of the nearest neighbors, whereas XGBoost is the gradient boosting model that is powerful, scalable, and has regularization in-built. The models were optimized using hyperparameters determined by initial experiments and cross-validation to perform better. Table III shows the most important hyperparameters of each baseline model.

TABLE III: Hyperparameters of Baseline Models

| Model | Hyperparameters |
| --- | --- |
| CatBoost | iterations=150, depth=6, learning_rate=0.1, l2_leaf_reg=3, random_seed=42, verbose=False |
| MLP | hidden_layer_sizes=(100, 50), activation=relu, solver=adam, alpha=0.0001, max_iter=300, random_state=42 |
| KNN | n_neighbors=7, weights=distance, algorithm=auto, leaf_size=30, p=2 (Euclidean) |
| XGBoost | n_estimators=150, max_depth=5, learning_rate=0.1, subsample=0.8, colsample_bytree=0.8, use_label_encoder=False, eval_metric=logloss, random_state=42 |

### G. Proposed StackLiverNet: A Stacking Ensemble Classifier

StackLiverNet is a stacking ensemble net which aimed to enhance the classification of liver disease by means of incorporating the predictive capability of numerous base classifiers. The basic versions are XGBoost, K-Nearest Neighbors (KNN). A meta-classifier is trained on their outputs as input features; this is done with LightGBM. Such a hierarchical method enables the meta-classifier to train in ways that it can combining the best of the base models in a manner that leads to better overall performance. Algorithm 2 demonstrates the process of stacking.

**Algorithm 2** StackLiverNet: Stacking Ensemble Algorithm

0: **Input:** Dataset $D = (X, y)$
0: Initialize base models: $M_1$ = XGBoost, $M_2$ = KNN
0: Initialize meta-model: $M_{meta}$ = LightGBM
0: Split $D$ into $k$ folds for cross-validation
0: **for** each fold $i = 1$ to $k$ **do**
0:    Train base models $M_1$ and $M_2$ on training folds $D \setminus D_i$
0:    Predict on validation fold $D_i$ using $M_1$ and $M_2$ to obtain predictions $P_1^{(i)}$ and $P_2^{(i)}$
0:    Concatenate predictions: $P^{(i)} = [P_1^{(i)}, P_2^{(i)}]$
0: **end for**
0: Combine all $P^{(i)}$ to form meta-training set $P$
0: Train meta-model $M_{meta}$ on $P$ with corresponding true labels $y$
0: **Output:** Final stacked model $\{M_1, M_2, M_{meta}\}$ =0

### H. Proposed Model Decision-Making Process

To improve interpretability, StackLiverNet integrates LIME for local explanations, SHAP for global feature impact, and the Morris method for sensitivity analysis. Together, these methods provide clear, complementary insights into the model's decision-making, enhancing transparency and clinical trust.

## IV. RESULTS AND DISCUSSION

Evaluating a model's performance across multiple metrics is essential to ensure its reliability and applicability in clinical settings. In this section, we present the results of StackLiverNet using accuracy, precision, recall, F1-score, Cohen's Kappa, and AUC to demonstrate its effectiveness in liver disease classification.

Table IV provides the summary of the comparative evaluation of the proposed StackLiverNet model with some well-known classifiers, such as CatBoost, MLP, KNN, and XGBoost, on training and testing datasets. The experiments prove that StackLiverNet outperforms other state-of-the-art methods on various metrics that are important in the tasks of classifying liver diseases. specifically, on the training set, StackLiverNet achieves an accuracy of 99.99%, a Cohens Kappa of 0.9998, and an F1-score of 0.9999, corresponding to an almost perfect accord and balanced predictive performance on both disease and non-disease classes. Notably, StackLiverNet achieves this high performance on the hidden testing set with an accuracy of 99.89%, a Kappa statistic of 0.9974 and an AUC of 0.9993. These findings indicate the high generalization capacity and stability of the model to distinguish correctly the cases of liver disease, reducing the false positive and the false negative. The value of Cohen s Kappa is high hence there is a strong agreement beyond chance which is very essential in medical diagnosis where reliability is of utmost importance. Relative to other baselines around KNN and XGBoost, though they too provide excellent classification scores, they just lose out on the overall performance that StackLiverNet provides, especially regarding Kappa and balanced F1-scores on the test dataset. Although the AUC values are high, CatBoost and MLP models demonstrate slightly lower Kappa scores, which may indicate the inability to effectively deal with the complex distribution of liver diseases or delicate interactions between clinical features. To further verify the robustness of the model and prevent overfitting we used 5-fold stratified cross-validation in the training and validation. This method maintains the ratio of liver disease and non-disease instances within each fold, and it gives a sensible estimate of model Performance in different data splits. The results of cross-validation are summarized in Table V.

Table V summarizes the mean accuracy and area under the curve (AUC) obtained through 5-fold stratified cross-validation for the evaluated models.Our suggested StackLiverNet model shows the best mean accuracy of 99.95% standard deviation being low, and mean AUC is 0.9996, which is also robust. This is not only a good predictive performance but also good stability across different data splits. Though other strong results like KNN and XGBoost would give similar results, StackLiverNet shows a slightly but significantly better result in terms of balancing accuracy and reliability. By contrast, CatBoost and MLP show rather low values of the mean, which may indicate their inability to adequately model the intricate patterns of liver disease data. These results strengthen the excellent generalization ability of StackLiverNet over the in-dependent test set assessment. To further prove the usefulness of these models in practical setting, we also examine their computational efficiency, namely the training and inference time, which is reported in Table V. These metrics are of great interest with regard to the feasibility assessment under real-world clinical deployment where timely decision support is needed.

TABLE IV: Performance Comparison of Models on Training and Testing Sets

| Model | Training | | | | | | Testing | | | | | |
|---|---|---|---|---|---|---|---|---|---|---|---|---|
| | Accuracy | Precision | Recall | F1-score | Kappa | AUC | Accuracy | Precision | Recall | F1-score | Kappa | AUC |
| CatBoost | 0.9849 | 0.9853 | 0.9849 | 0.9849 | 0.9697 | 0.9997 | 0.9679 | 0.9497 | 0.9773 | 0.9620 | 0.9240 | 0.9991 |
| MLP | 0.9794 | 0.9802 | 0.9794 | 0.9794 | 0.9587 | 0.9996 | 0.9643 | 0.9447 | 0.9747 | 0.9578 | 0.9157 | 0.9976 |
| KNN | **0.9999** | 0.9999 | 0.9999 | 0.9999 | **0.9998** | 0.9999 | 0.9981 | 0.9969 | 0.9984 | 0.9976 | 0.9953 | 0.9992 |
| XGBoost | 0.9991 | 0.9991 | 0.9991 | 0.9991 | 0.9981 | 0.9999 | 0.9974 | 0.9958 | 0.9980 | 0.9969 | 0.9937 | 0.9995 |
| StackLiverNet | **0.9999** | 0.9999 | 0.9999 | 0.9999 | **0.9998** | 0.9998 | **0.9989** | 0.9986 | 0.9988 | 0.9987 | **0.9974** | 0.9993 |

TABLE V: Mean 5-Fold Cross-Validation Accuracy and AUC for Liver Disease Detection Models

| Model | Mean Accuracy | Mean AUC |
|---|---|---|
| CatBoost | 0.9708 ± 0.0046 | 0.9992 ± 0.0002 |
| MLP | 0.9785 ± 0.0106 | 0.9978 ± 0.0008 |
| KNN | 0.9993 ± 0.0004 | 0.9996 ± 0.0004 |
| XGBoost | 0.9986 ± 0.0006 | 0.9998 ± 0.0003 |
| **StackLiverNet** | **0.9995 ± 0.0004** | **0.9996 ± 0.0004** |

The training and inference time of the considered models is summarized in Table VI. Multi-Layer Perceptron (MLP) has the longest training time of about 36 seconds, which indicates the computing requirement of neural network training. On the contrary, K-Nearest Neighbors (KNN) and CatBoost models train significantly faster, and KNN takes approximately 0.015 seconds. In terms of inference, CatBoost is the fastest in prediction with 0.0065 seconds. The only disadvantage is that StackLiverNet requires a little more time to train and make an inference, which is around 4.28 seconds and 0.11 seconds, respectively, but these gaps are insignificant, and in clinical practice, fractions of a second do not impact diagnostic workflow and decision-making performance. The computational time advantage of StackLiverNet margins is offset by its classification accuracy advantages, making it adequate to be deployed in real-world applications in the detection of liver diseases.

TABLE VI: Training and Inference Times for Liver Disease Classification Models

| Model | Training Time (seconds) | Inference Time (seconds) |
|---|---|---|
| CatBoost | 0.3897 | 0.0065 |
| MLP | 35.9576 | 0.0479 |
| KNN | 0.0151 | 0.0447 |
| XGBoost | 0.1802 | 0.0268 |
| StackLiverNet | 4.2783 | 0.1106 |

*A. Performance Visualization of the Proposed StackLiverNet*

In order to prove the efficiency of the proposed StackLiverNet model, some performance visualization methods were utilized, such as a confusion matrix, 5-fold cross-validation plots, and ROC curve analysis, as presented in Figure 3. The concept of the confusion matrix shows clearly that the model has the capability of separating the classes with high precision. On the total samples, the model managed to identify 3,332 negative and 1,337 positive cases correctly with only 3 negative and 2 positive samples being misclassified. It is an extremely low error rate and it shows the high predictive abilities of the model. The model robustness and generalizability are also proved by the 5-fold cross-validation results. The accuracy and AUC scores were similarly high (above 0.999) and heavily stable throughout all folds. This stability indicates that Stack-LiverNet generalizes well on various data subsets and it does not overfit. Besides, ROC curve indicates that the model has an excellent discriminative power. This curve almost borders the upper-left point of the plot and the computed value of AUC of 0.9993 supports the almost perfect behavior of the classifier to distinguish between the two classes. Generally, these visual findings prove that StackLiverNet provides high precise, stable, and generalizable predictions, and it is a robust model to detect liver disease.

*B. Explainability of the Proposed StackLiverNet Model*

In order to improve the interpretability of the suggested StackLiverNet model, the Local Interpretable Model-agnostic Explanations (LIME) method was applied to examine single predictions of each of the two classes. Figure 4 shows the LIME explanations of the representative samples of Class 0 (liver disease presence) and Class 1 (liver disease absence). In the Class 0 sample (liver disease) (Figure 4, top), the model put great emphasis on the high level of *Alkaline Phosphatase (Alkphos ¿ 0.28)* and moderate range of *SGOT (Aspartate Aminotransferase)* to back the decision. Factors like low *SGPT (Alamine Aminotransferase)*, *Total Bilirubin* and *Albumin* had non-positive effect on this decision meaning that they slightly pulled the prediction towards Class 1 (no liver disease) but were overridden by the positive evidence. On the other hand, Class 1 sample (no liver disease) (Figure 4, bottom) was predominantly shaped by higher *Albumin (ALB ¿ 0.84)* and lower *Total Bilirubin* and *SGPT*. These aspects lent great positive evidence towards the Class 1 prediction. Conversely, higher concentration of *Alkaline Phosphatase* and *SGOT* had negative effects, as it slightly altered the prediction toward Class 0 (liver disease).

These LIME explanations allow seeing the model reasoning process and depicting its dependence on biologically rele-vant features to classify liver disease. The local explanations provided by LIME boost the confidence in the predictions carried out by StackLiverNet and enforces its diagnosability in a clinical decision-making process.

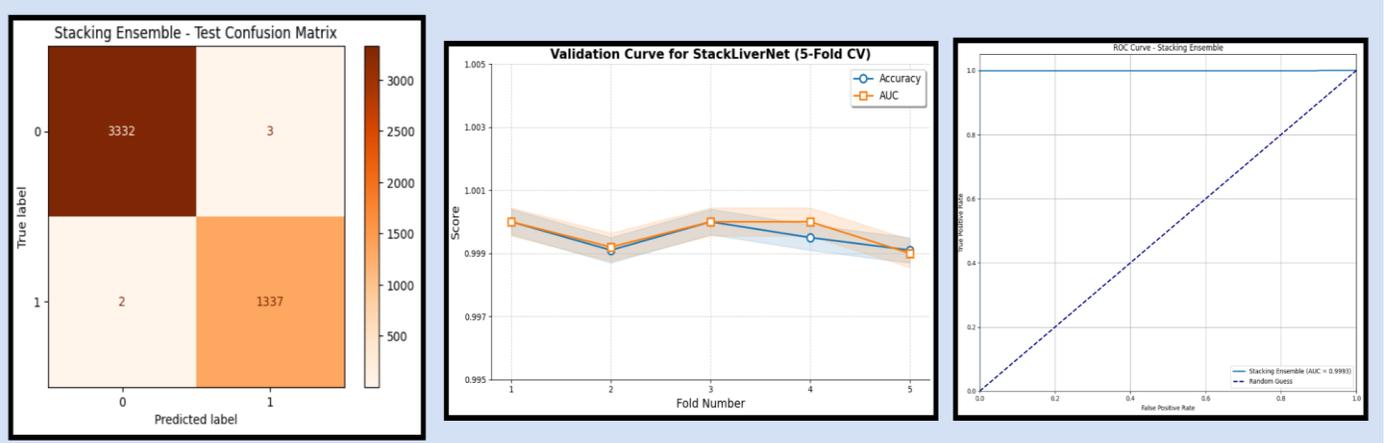

Fig. 3: Comprehensive evaluation results of the StackLiverNet model: confusion matrix (eft), Validation Curve (Middle), ROC Curve (Right).

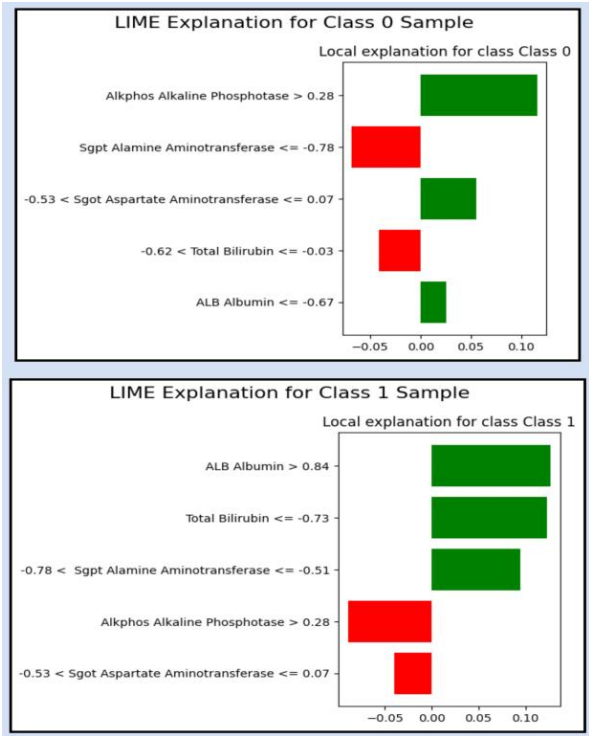

Fig. 4: LIME explanations showing the most influential features for Class 0 (top) and Class 1 (bottom) predictions. Green bars indicate positive contribution towards the predicted class, while red bars indicate negative influence.

Additionally, SHAP (SHapley Additive exPlanations) was employed to provide global interpretability, ranking features by their average impact on model output. SHAP confirms the biological validity of the model's decisions by highlighting important biomarkers such as Total Bilirubin, Alkaline Phosphotase, and SGPT. Together, LIME and SHAP enhance the interpretability and diagnosability of StackLiverNet, offering both local and global insights into its predictions.

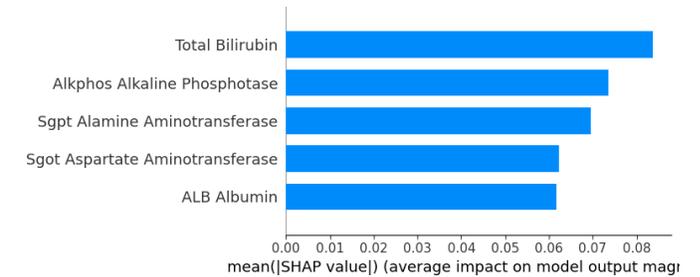

Fig. 5: SHAP summary plot showing global feature importance across the dataset. The x-axis represents the mean absolute SHAP value, indicating the average contribution of each feature to the model's predictions.

To complement SHAP and LIME, we applied the Morris method, a global sensitivity analysis technique that estimates the mean absolute effect ($\mu*$) and variability ($\sigma$) of each feature on model output. As shown in Table VII, **ALP** (Alkaline Phosphotase) and **SGPT** (Serum Glutamate Pyruvate Transaminase) had the highest $\mu*$ values, indicating strong influence. In contrast, **SGOT** (Serum Glutamate Oxaloacetate Transaminase) and **ALB** (Albumin) showed minimal effect, consistent with SHAP results and confirming limited predictive relevance in this setting.

### C. Comparative Analysis with Previous studies

Table VIII provides the comparison of different machine learning and deep learning models used to classify liver disease using Liver Patient Dataset (LDPD). The majority of research adopted boosting or ensemble-based methods to enhance the accuracy, and the recorded results were in the range of

TABLE VII: Morris Sensitivity Analysis with short feature names. $\mu$ = mean effect, $\mu*$ = absolute mean effect, $\sigma$ = standard deviation, and $\mu*_{conf}$ = confidence interval.

| Feature | $\mu$ | $\mu*$ | $\sigma$ | $\mu*_{conf}$ |
|---|---|---|---|---|
| TB | -0.0059 | 0.0060 | 0.0089 | 0.0048 |
| ALP | -0.2942 | 0.2942 | 0.6130 | 0.3669 |
| SGPT | -0.1441 | 0.1442 | 0.4522 | 0.2822 |
| SGOT | 0.0008 | 0.0008 | 0.0024 | 0.0014 |
| ALB | -0.0000 | 0.0000 | 0.0001 | 0.0000 |

98.73% to 99.8%. remarkably, our suggested StackLiverNet framework was in a position to attain an almost optimal accuracy of 99.89%, surpassing various recent machine learning approaches, and preserving interpretability and efficiency. StackLiverNet, in contrast to most of the previous models, includes extensive preprocessing and explainable AI, being both high-performance and clinically transparent.

TABLE VIII: Comparison of Liver Disease Classification Models

| Study | Dataset | Model | Accuracy (%) |
|---|---|---|---|
| [5] | LDPD | Gradient Boosting | 98.80 |
| [6] | LDPD | CNN-LSTM | 98.73 |
| [7] | LDPD | Stacked RF | 99.72 |
| [8] | LDPD | Boosting | 98.80 |
| [9] | LDPD | EBM | $\approx$ 99.78 |
| **Our Study** | LDPD | StackLiverNet | **99.89** |

## V. CONCLUSION

In this work, we introduced StackLiverNet, a stacked ensemble framework for liver disease classification, which achieved an impressive 99.89% accuracy on the test set with just 5 misclassifications. The framework offers robust, consistent and reliable predictions because of a suitable combination of sophisticated data preprocessing mechanisms, recursive feature selection, and class imbalance treatment procedures with a significantly expressive LightGBM meta-model. Besides that, the Local Interpretable ModelAnagnostic Explanations (LIME) capability has very good interpretability and selected key clinical features like, Alkaline Phosphatase and SGOT, which also marks in line with medical knowledge. SHapley Additive exPlanations (SHAP) was used as complementary to LIME to achieve global interpretability; in particular, SHAP ranked the features by average effect on the model output, validating the discovery of important biomarkers Total Bilirubin, Alkaline Phosphatase, and SGPT. In an additional support to the interpretability, the Morris method, a global sensitivity analysis approach was used, where the mean absolute effect and variability of each feature as input variable was calculated on the model predictions. The outcome also revealed that the greatest impact was exerted by the two features, namely ALP and SGPT, and the least impact was exerted by the two features, namely SGOT and Albumin, which were confirmed by SHAP results. Such accuracy and multidimensional understanding also give StackLiverNet a high viability and reliability rate as an instrument that can help healthcare professionals in the diagnosis of liver diseases at an early stage and offers possible treatment in time.

Future work will focus on the validation of the model on larger heterogeneous hospital data in order to determine its generalizability and clinical utility. Implementing StackLiverNet into real working conditions of healthcare facilities will help to test how the technology is integrated and what impact it produces on patients. Additionally, integration of multimodal data such as imaging and strengthening of explainability methods will further increased the functioning of the models and confidence of the clinicians.